\documentclass[10pt,journal,compsoc]{IEEEtran}
\ifCLASSOPTIONcompsoc
  \usepackage[nocompress]{cite}
\else
  \usepackage{cite}
\fi
\ifCLASSINFOpdf
\else
\fi

\usepackage{tabularx}
\usepackage{siunitx}
\usepackage{caption}
\usepackage{bm}
\usepackage{subfigure}
\usepackage{enumerate}
\usepackage{arydshln}
\usepackage{threeparttable}
\usepackage{hhline}
\usepackage{algorithmic}
\usepackage{algorithm}
\usepackage{color}
\usepackage{xfrac}
\usepackage{amsthm}
\usepackage{booktabs}
\usepackage{multirow}
\usepackage{graphicx}
\usepackage{animate}

\definecolor{myGreen}{RGB}{46,204,113}
\definecolor{myRed}{RGB}{255,0,0}
\definecolor{myYellow}{RGB}{255,255,0}
\definecolor{myBlue}{RGB}{52,152,219}

\newcolumntype{C}{>{\centering\arraybackslash}X}

\usepackage[breaklinks=true]{hyperref}

\begin{document}
%
\title{Stereo Unstructured Magnification: Multiple Homography Image for View Synthesis}
%
%
%
%

\author{Qi~Zhang*,
        Xin~Huang*,
        Ying~Feng,
        Xue~Wang,
        Hongdong~Li
        and~Qing~Wang
\IEEEcompsocitemizethanks{\IEEEcompsocthanksitem Qi Zhang, Xin Huang, Xue Wang and Qing Wang (corresponding author) are with the School of Computer Science, Northwestern Polytechnical University, Xi'an 710072, China (e-mail: qwang@nwpu.edu.cn).
\IEEEcompsocthanksitem Ying Feng is with Tencent AI Lab, Shenzhen 518054, China.
\IEEEcompsocthanksitem Hongdong Li is with the ANU and ACRV, Australian National University, Australia (e-mail: hongdong.li@anu.edu.au).
\protect\\
\IEEEcompsocthanksitem Qi Zhang and Xin Huang contributed equally to this work.
\IEEEcompsocthanksitem The work was supported by NSFC under Grant 62031023, 61801396.}
\thanks{Manuscript received April 19, 2005; revised August 26, 2015.}}

%
%

\markboth{Journal of \LaTeX\ Class Files,~Vol.~14, No.~8, August~2015}%
{Zhang \MakeLowercase{\textit{et al.}}: Multiple Homography Image for View Synthesis}
%



\IEEEtitleabstractindextext{%
\begin{abstract}
This paper studies the problem of view synthesis with certain amount of rotations from a pair of images, what we called stereo unstructured magnification. While the multi-plane image representation is well suited for view synthesis with depth invariant, how to generalize it to unstructured views remains a significant challenge. This is primarily due to the depth-dependency caused by camera frontal parallel representation.  Here we propose a novel multiple homography image (MHI) representation, comprising of a set of scene planes with fixed normals and distances. A two-stage network is developed for novel view synthesis. Stage-1 is an MHI reconstruction module that predicts the MHIs and composites layered multi-normal images along the normal direction. Stage-2 is a normal-blending module to find blending weights. We also derive an angle-based cost to guide the blending of multi-normal images by exploiting per-normal geometry. Compared with the state-of-the-art methods, our method achieves superior performance for view synthesis qualitatively and quantitatively, especially for cases when the cameras undergo rotations.
\end{abstract}

\begin{IEEEkeywords}
View Synthesis, Scene Representation, Multiple Homography Image, Homography Transformation.
\end{IEEEkeywords}}

\maketitle

\IEEEdisplaynontitleabstractindextext

%
\IEEEpeerreviewmaketitle

\IEEEraisesectionheading{\section{Introduction}\label{sec:introduction}}

\begin{figure*}[tb]
    \centering
    \includegraphics[width=\hsize]{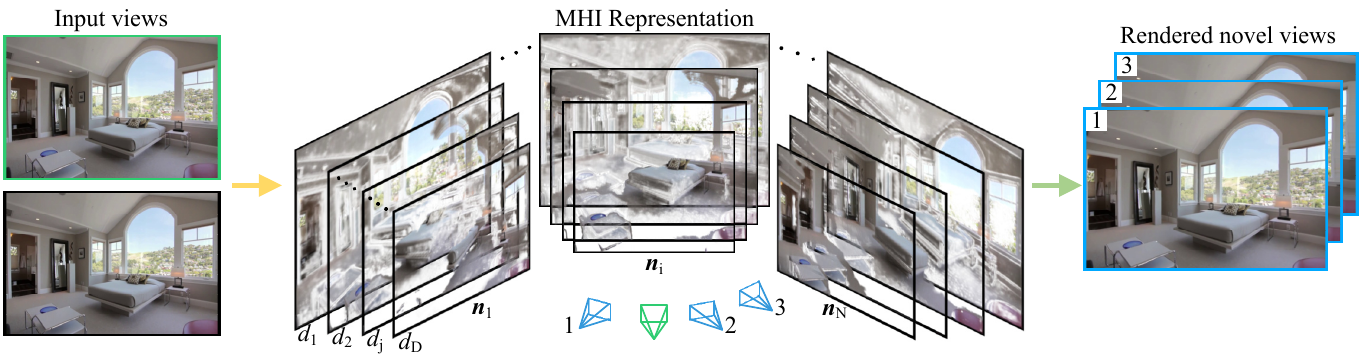}
    \caption{
    We predict the multiple homography image (MHI) representation of the scene in the reference view ({\color{myGreen} green}) from two input images and render unstructured views ({\color{myBlue} blue}) with a certain amount of camera rotations via a two-stage network.}
    \label{fig:overview}
\end{figure*}

\IEEEPARstart{L}{ight-field} rendering provides an effective way to endow the users with a compelling immersive visual experience to view the scene from different vantage points or novel viewing orientations \cite{levoy1996light,gortler1996lumigraph}.  Technically, this requires the capability of synthesizing novel view images from a pre-captured sparse set of input multi-view images. Recent advent of deep learning techniques has offered a powerful method to solve such a novel view synthesis problem.  However, previously published methods \cite{kalantari2016learning,wu2017light,li2020synthesizing} are often restricted to cases requiring all cameras are parallel or regularly arranged.  It remains a challenging task to use stereo images to synthesize novel views with certain amount of rotations 
(\textit{i.e.}  so-called ``stereo unstructured magnification'').

The multi-plane image (MPI) representation (due to \textit{e.g.} \cite{szeliski1998stereo, szeliski1999stereo, zhou2018stereo, mildenhall2019local, flynn2019deepview, srinivasan2019pushing})
recovers a set of depth-dependent images with transparency and is well-suited to rendering by convolutional neural networks.
Such MPI representations only partition the scene into multiple depth-discretized frontal-parallel planes, while its normal aligns with the camera's principal axis. Similarly, the layered depth image (LDI) \cite{shade1998layered, tulsiani2018layer,choi2019extreme} is another depth-dependent representation which cannot handle complex appearance effects well. However, in real world scenarios, scene planes are not necessarily parallel to the camera plane. 

These depth-based representations tend to divide a scene plane into multiple sub-planes based on different depth layers and be sensitive to slight rotations. As a result, ghosting artifacts and duplicate edges may occur in the rendered views, especially on slanted surfaces. Moreover, the spatial structure consistency of camera-parallel planes is hard to maintain during unstructured view synthesis.
 
Neural radiance fields (NeRF) \cite{mildenhall2020nerf} recover a continuous 5D volumetric function to represent the mapping between the ray coordinates and the radiance fields of rays for view synthesis. Despite its astonishing results, NeRF has no generalization to other scenes, and rendering is a time-consuming procedure.
Besides, stable view synthesis (SVS) \cite{riegler2021stable} encodes the 3D scene structure with compact directional feature vectors for view synthesis, but the quality and runtime are limited by the traditional multi-view reconstruction methods that pre-compute the 3D scene structure. Unlike neural implicit representations that consider dense inputs, we only have two views as input to render novel views.


Our goal is to learn a deep neural network to infer a layered scene representation suitable for rendering unstructured views from two images, and in particular extrapolating views with a certain amount of camera rotations.
To handle challenging cases like slanted surfaces and rotated view rendering, we introduce homography instead of depth to layer scenes into planes (so-called ``multiple homography image''). With the addition of normal, given that the alpha composition is used for blending depth-based layers, the per-normal geometry and the corresponding blending method are proposed to predict the weights for compositing homography-based layers. 
The overview of the proposed method is shown in Fig. \ref{fig:overview}. 
In experiments, the proposed scene presentation is compared with recent MPI representation, and performs a number of ablation studies. Our representation outperforms the MPI-based method for view synthesis qualitatively and quantitatively, especially in cases with rotated cameras and slanted scene planes.

In summary, the paper's specific contributions are:

\begin{enumerate}
    \item A \textit{multiple homography image} (MHI) representation that contains a set of scene planes with fixed normals and distances is proposed (Sec. \ref{subsec:MHI}).
    \item A two-stage learning-based framework for stereo unstructured magnification, including an MHI reconstruction module and a normal-blending module, is developed (Sec. \ref{sec:pipeline}).
    \item The per-normal geometry is explored to guide the normal-blending module to predict weights of multi-normal images (Sec. \ref{subsec:normal_blending}).
\end{enumerate}

\section{Related works}
Our goal is to seek a scene representation for unstructured view synthesis from stereo images. We are inspired by several previous methods in view synthesis, scene representation and image blending.
\subsection{View Synthesis}
View synthesis is a complicated issue that has sparked a lot of interest fields in computer vision and graphics. Earlier lines of researches could be considered as an image-based warping task combined with geometry structure~\cite{shum2000review}, such as implicit geometry from dense images (so-called ``light-field rendering'')~\cite{mcmillan1995plenoptic, levoy1996light, gortler1996lumigraph, buehler2001unstructured, davis2012unstructured}, depth and planar homography~\cite{zitnick2004high, chaurasia2013depth, zhou2013plane, sinha2014efficient, cayon2015bayesian, overbeck2018system}, and 3D explicit geometry~\cite{debevec1996modeling, yu20143d, hedman2017casual, hedman2018instant}. In particular, 
Debevec \textit{et al.} \cite{debevec1996modeling} present an influential view-dependent texture mapping method to render novel views via blending nearby captured views that have been projected using a global mesh.
Unfortunately, estimating high-quality meshes with geometric boundaries that fit well with edge images is challenging. Chaurasia \textit{et al.} \cite{chaurasia2013depth} use the superpixel as a local geometric element to warp the novel view and blend the mapped views with weights specified by camera orientation and the reliability of depth information. It uses local geometry instead of difficult and expensive global mesh reconstruction. However, it relies heavily on the estimated depth, which is sensitive to occluded regions. Hedman \textit{et al.} \cite{hedman2018instant} propose a pipeline that includes both global mesh and local geometry, but the mesh reconstruction task is time-consuming.
In summary, these traditional image-based methods cannot handle the scene with complex appearance effects and suffer from significant artifacts.

In recent years, deep learning techniques have been applied in many approaches to perform novel view synthesis.
Previous learning-based methods can be classified into three categories according to the number of input images. One category tries to learn the scene structure and synthesize novel views from such limited input as a single image~\cite{zhou2016view, liu2018geometry, niklaus20193d, wiles2020synsin, hu2021worldsheet, shih20203d}. Considering the lack of geometric constraints, it is hard for these methods to synthesize visually acceptable views and preserve the scene structure. Other learning-based methods focus on the challenging problem of training networks to learn about arbitrarily-distant views combined with scene structure from multiple inputs~\cite{hedman2018deep, xu2019deep, riegler2020free, shi2021self, riegler2021stable}. 
However, these methods need off-line structure-from-motion algorithms to calculate geometric structure from multiple images for supervision. Another exciting advance is neural radiance fields (NeRF) \cite{mildenhall2020nerf}, which optimizes a mapping from continuous 3D coordinates and 2D view directions to a 4D radiance including RGB values and volume density. It has many variants, such as NeRF for wild images \cite{martin2021nerf}, MIP-NeRF \cite{barron2021mip}, NeRF for dynamic scenes \cite{li2021neural, pumarola2021d, xian2021space} and NeRF for generation \cite{schwarz2020graf, trevithick2021grf, niemeyer2021giraffe}. Although NeRF-based methods have shown promising view synthesis results, they need to overfit to the given scene with enough samples, using time-consuming per-scene training.
Besides, our approach belongs to the category of view synthesis from stereo images~\cite{zhang2015light, flynn2016deepstereo, zhou2018stereo, choi2019extreme}. It predicts layered images combined with depth information to synthesize views for specific applications, such as planar light-field rendering and baseline magnification.

\subsection{Scene Representation}
The most relevant methods to our approach are algorithms that predict a 3D scene representation from stereo images, and render novel views using the representation of the reference image via differentiable parallel projection. Tulsiani \textit{et al.} \cite{tulsiani2018layer} propose a learning-based method to infer the LDI representation \cite{shade1998layered}, which is a generalization of depth. It is not sensitive to complex appearance effects, \textit{e.g.} semi-transparency and non-Lambertian. Penner and Zhang \cite{penner2017soft} compute a local layered softness image for scene representation and blend these volumes to render novel views.
Recently, Zhou \textit{et al.} \cite{zhou2018stereo} propose the MPI representation to represent the scene using several layers of color images and alpha transparencies at different depths \cite{szeliski1998stereo, szeliski1999stereo}, where novel views are forward projected and alpha blended from MPIs. MPI representation has been applied in many scenarios, such as extending baselines \cite{zhou2018stereo}, predicting occluded contents for view synthesis \cite{srinivasan2019pushing}, blending multi-view MPIs in the rendered view \cite{mildenhall2019local}, synthesizing views with learned gradient descent \cite{flynn2019deepview}, rendering views from a single image \cite{tucker2020single, li2020synthesizing}, synthesizing views of dynamic scenes \cite{lin2021deep} and variants as multi-sphere image \cite{broxton2020immersive, attal2020matryodshka}.

These representations discretize the scene with multiple camera-parallel planes at fixed depths, where their normals are defined by the camera orientation. However, in real world scenarios, many scene planes cannot be equivalent to such camera-parallel planes, especially for the scene plane with a large slope. It may divide the scene plane into different layers, which will cause ghosting artifacts and seams and destroy the spatial structure consistency on the rendered views. Slight camera rotations will aggravate such conditions. In contrast, the proposed MHI representation utilizes approximate scene planes (normal and distance) instead of the camera-parallel planes (depth) to achieve scene discretization for unstructured view synthesis, so the MHI could be thought of as a generalization of MPI representation.

\subsection{Image Blending}
Blending multiple layered images is crucial to compensate for geometric and visibility errors, and for the layer-dependent and view-dependent effects. A variety of works for view synthesis has used image blending to heuristically achieve soft visibility, including discretized soft visibility \cite{penner2017soft}, alpha blending of the MPIs \cite{zhou2018stereo, flynn2019deepview, srinivasan2019pushing}, blending multi-view images in the target view \cite{hedman2018deep, riegler2020free, riegler2021stable} and accumulated alpha blending of multi-view MPIs \cite{mildenhall2019local}. In particular, Buehler \textit{et al.} \cite{buehler2001unstructured} derive the per-pixel blending weights for images projected by a global mesh, then propose a heuristic algorithm for unstructured light-field rendering.
Mildenhall \textit{et al.} \cite{mildenhall2019local} use the distance between translation vectors of poses to compute soft blending weights for a weighted combination of renderings from MPIs of different input images. Hedman \textit{et al.} \cite{hedman2018deep} present a neural network to learn the blending weights that are then used to combine projected contributions from multiple nearby images.
Correctly choosing appropriate weights for specific inputs in the presence of insufficient geometric information is still a challenging problem which has plagued the field from the outset. In this paper, we explore the per-normal geometry and compute an angle-based cost to find approximate scene planes that are used to supervise normal blending weights for each pixel.

\section{Scene Representation}\label{sec:representation}
\subsection{Multiple Homography Image Representation}\label{subsec:MHI}

As we known, the scene plane is usually defined in a general form $\bm{n}^\top\bm{X}\!+\!d\!=\!0$, where $\bm{n}$ and $d$ are the normal of the plane and the distance between the plane and the camera center. The core of our scene representation for view synthesis pipeline is composed of a collection of scene planes, where each plane has a constant normal and distance (so-called homography),
while the MPI \cite{zhou2018stereo} is the special case of the MHI that fixes the normals perpendicular to the camera plane.
Unlike MPI only relating to depth, whose normal is constant and perpendicular to the camera plane, the proposed MHI can be parameterized by the normal $\bm{n}$ and the distance $d$ of the scene plane. One advantage of such representation is that it is equivalent to generalizing MPIs with different normals, achieving unstructured view synthesis without ghosting artifacts and structure inconsistency. 

In summary, \textit{multiple homography images} (MHIs) are defined as a set of images projected from the scene planes at a fixed range of normals and distances to achieve scene discretization, where each scene plane $(\bm{n}^\top,d)^\top$ decodes a color image $C$ and a visible map $\alpha$. 
MHI representation is therefore described as a sequence of such layers $\{(C_{11}, \alpha_{11}), \cdots, (C_{N\!D}, \alpha_{N\!D})\}$, where $N$ and $D$ are the number of normals and the number of distances for each normal respectively, as shown in Fig. \ref{fig:overview}. With the introduction of normals, MHIs represent the scene geometry and texture more flexibly, especially for planes with large slopes. Also, adding samples of normals to scene representation allow us to discretize the scene more finely and apply to unstructured view synthesis with a range of camera rotations.

\subsection{Multi-Normal Image}\label{subsec:MNI}
Suppose we have predicted MHIs of the scene in the reference image $I_r$, we map MHIs into the target camera via the inverse planar transformation~\cite{hartley2003multiple}. Given intrinsic matrices of the reference and target camera denoted as $\bm{K}_r$ and $\bm{K}_t$ respectively, and the relative pose from the target camera to the reference camera defined with rotation $\bm{R}$ and translation $\bm{t}$, we then formulate the differentiable homography of MHIs,
\begin{equation}
	\left[\begin{array}{c}
		u_r \\ v_r \\ 1
	\end{array}	\right]\sim
	\bm{K}_r(\bm{R}-\frac{\bm{t}\bm{n}^\top\bm{R}}{d+\bm{n}^\top\bm{t}})\bm{K}_t^{-1}
	\left[\begin{array}{c}
		u_t \\ v_t \\ 1
	\end{array}	\right],
	\label{eq:diff_homography}
\end{equation}
where $(u_r, v_r, 1)^\top$ and $(u_t, v_t, 1)^\top$ are corresponding pixels in the reference and target MHIs. Note that, $(\bm{n}^\top, d)^\top$ represents an MHI plane in the reference view. Therefore, layered color image $\hat{C}_{ij}$ and visible map $\hat{\alpha}_{ij}$ of the target MHI can be obtained from the reference MHI $(C_{ij}, \alpha_{ij})$ according to Eq. \ref{eq:diff_homography}. 
The reason for using backward projection is to eliminate irregular sampling via bilinear interpolation.

MHIs with the same normal are parallel and satisfied with the back-to-front order, and the visible map indicates the visibility of the MHI plane that helps us to softly handle edges and occlusions. Similar to alpha blending of MPIs, partial pixels in the target image could also be synthesized from MHIs with the same normal via standard over-composition \cite{porter1984compositing, szeliski1998stereo, szeliski1999stereo}. They thus form a normal layer which we call the \textit{multi-normal image}. Specifically, based on the transformed visible map $\{\hat{\alpha}_{i1},\cdots, \hat{\alpha}_{iD}\}$, the multi-normal image $\hat{C}_i$ can be computed via blending the color images $\{\hat{C}_{i1},\cdots,\hat{C}_{iD}\}$ in back-to-front order,
\begin{equation}
	\hat{C}_i=\sum_{j=1}^{D}{\hat{C}_{(i,j)}\underbrace{\hat{\alpha}_{(i,j)}\prod_{k=1}^{j-1}\left(1-\hat{\alpha}_{(i,k)}\right)}_{\text{per-homography opacities}}}.
	\label{eq:normal_alpha_blend}
\end{equation}

\begin{figure}[tb]
	\includegraphics[width=\hsize]{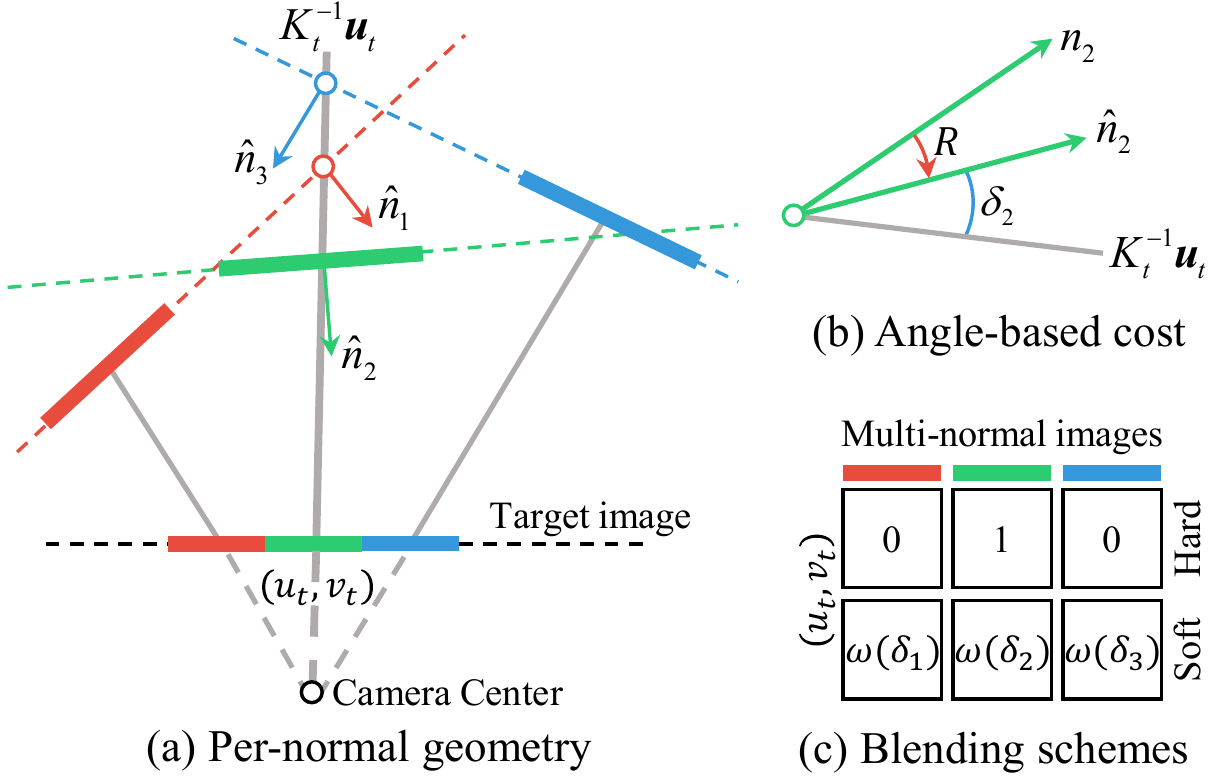}
	\caption{Blending of multi-normal images. (a) shows the per-normal geometry to find appropriate normals for each pixel.
	(b) indicates an angle-based cost between the normal and pixel.
	(c) compares the hard and soft blending schemes.
	}
	\label{fig:MNI_blending}
\end{figure}

\begin{figure*}[tb]
	\centering
	\includegraphics[width = \hsize]{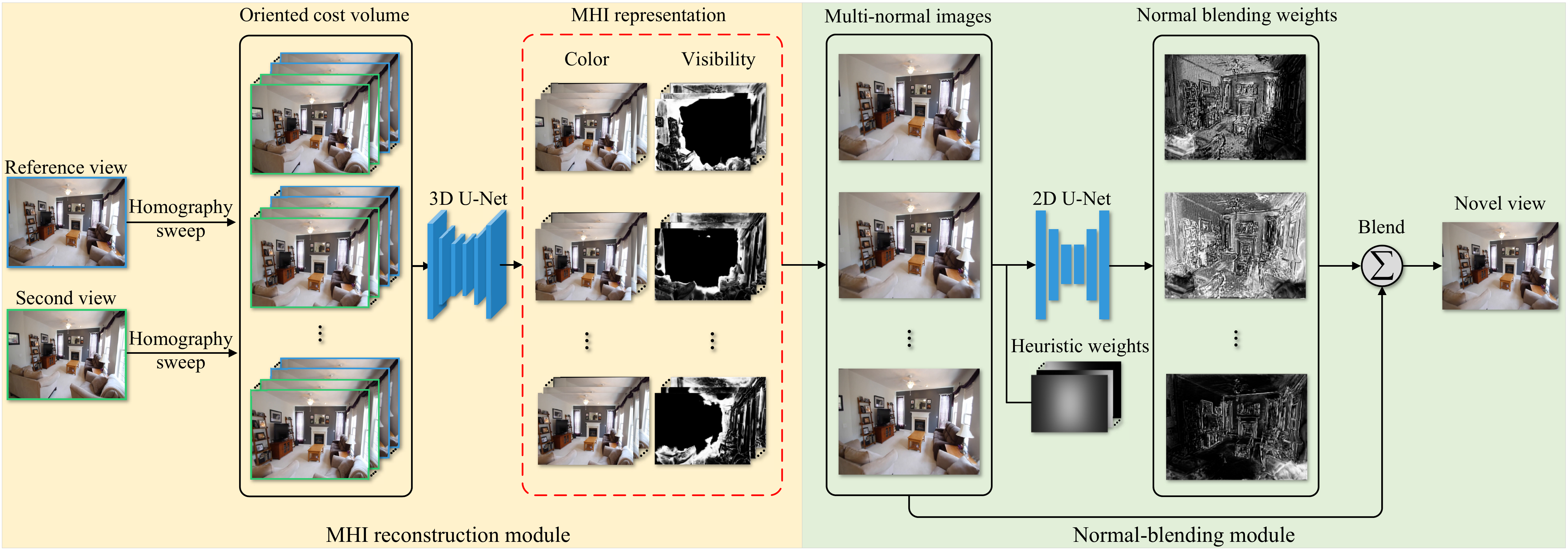}
	\caption{The pipeline of the proposed two-stage network. The whole network architecture consists of two stages: MHI reconstruction module and normal-blending module.}
	\label{fig:network}
\end{figure*}

\subsection{Per-Normal Geometry}\label{subsec:normal_blending}
The multi-normal images denote alpha compositing images of MHIs along different normal directions. 
However, a single multi-normal image alone will not necessarily render all continue planes due to the normal differences between the multi-normal image and scene plane. 
Therefore, once we have composited multi-normal images $\{\hat{C}_1,\cdots,\hat{C}_N\}$ in the target view from mapped MHIs, it is crucial to choose appropriate blending weights for these images, which captures scene-dependent effects while minimizing ghosting artifacts and preserving structure consistency, especially for scene planes having large slopes and rich textures.

Fig. \ref{fig:MNI_blending} illustrates the multi-normal images based blending for novel view synthesis. We first explore the per-normal geometry to compute a per-pixel cost for each multi-normal image. Here, we assume that the ray captured by the target camera should be as perpendicular to the scene plane as possible. As shown in Fig. \ref{fig:MNI_blending}(a), pixel $(u_t,v_t)^\top$ in the target image is equivalent to a single ray emitted from different scene planes and may appear in different multi-normal images. Normal blending is a process that finds an approximate normal for each pixel from multi-normal images. Based on the perpendicular assumption, the angle between the ray and the normal of the multi-normal image helps to preserve plane continuity and reproduce normal-dependent effects, namely scene planes are approximately facing towards the center of the target camera. Similar to \cite{buehler2001unstructured}, we therefore derive an angle-base cost that prioritizes which multi-normal images to be blended for every pixel in the novel view,
\begin{equation}
	\delta\left(\bm{n}_i, \bm{u}_t\right)=\frac{\left|\left(\bm{R}^\top\bm{n}_i\right)^\top\bm{K}_t^{-1}\bm{u}_t\right|}{\left\|\bm{R}^\top\bm{n}_i\right\|_2\left\|\bm{K}_t^{-1}\bm{u}_t\right\|_2},
	\label{eq:angle_cost}
\end{equation}
where $\bm{n}_i$ is the normal of $i$-th multi-normal image in the reference view, $\bm{u}_t$ is homogeneous image coordinates in the target view. As shown in Fig. \ref{fig:MNI_blending}(b), Eq. \ref{eq:angle_cost} indicates cosine of the angle between layered normal $\hat{\bm{n}}_i$ and ray direction $\bm{u}_t$ in the target view, where $\hat{\bm{n}}_i$ is rotated from layered normal ${\bm{n}}_i$ in the reference view, \textit{i.e.} $\hat{\bm{n}}_i=\bm{R}^\top\bm{n}_i$.
It is interesting to note that the translation is neglect during normal blending. The reasons are: the normal direction is constant with a translation, and the influence of translation is already considered in the alpha blending of the multi-normal image.

According to Eq.~\ref{eq:angle_cost}, two blending schemes, including hard blending and soft blending, are proposed to adaptively blend the multi-normal images as shown in Fig. \ref{fig:MNI_blending}(c). We use normal-dependent blending weights $\omega_i$, each formulated by the angle-based cost $\delta_i$ via hard or soft blending scheme and normalized so that the resulting composited image $\hat{I}_t$ is fully opaque,
\begin{equation}
	\hat{I}_t=\frac{\sum_{i}^{N}{\omega_i\hat{C}_i}}{\sum_{i}^{N}{\omega_i}}.
	\label{eq:MNI_blending}
\end{equation}
In hard blending scheme, the weight $\omega_i$ of the multi-normal image which has the largest angle-based cost is set to one, whereas the rest weights to zeros, \textit{i.e.} logical weight, as shown in the first row of Fig. \ref{fig:MNI_blending}(c). However, the hard blending is a simple cut-to-pasting, which could cause misalignment and harshness on the boundaries of weights. The soft blending scheme is therefore proposed. We use the weight $\omega_i$ sampled from an exponential function, namely $\omega(\delta_i)\!=\!e^{3(2\delta_i-1)}$, as shown in the second row of Fig. \ref{fig:MNI_blending}(c). The soft blending scheme maintains structure consistency but generates ghosting and aliasing artifacts. Consequently, to combine the two schemes' advantages, we consider a network guided by per-normal geometry to learn blending weights. 

\section{Deep View Synthesis Pipeline}\label{sec:pipeline}
The key of the proposed method is to perceive MHIs and blend multi-normal images via a deep learning framework for synthesizing a target image $I_t$ of the same scene from the reference image $I_r$ with the supervision of the second image $I_s$. The camera parameters of input images and output image are also provided, where extrinsic parameters are relative pose with respect to the reference camera. 
As shown in Fig. \ref{fig:network}, the proposed network mainly consists of an MHI reconstruction module that extracts the MHIs from inputs and a normal-blending module which renders novel views via the blending of alpha composited multi-normal images. 

\subsection{MHI Reconstruction Module} \label{subsec:MHI-module}
Inspired by the recent advances of MPI reconstruction \cite{zhou2018stereo,srinivasan2019pushing,mildenhall2019local}, we use a 3D U-Net-based \cite{cciccek20163d} convolutional architecture $\phi_{R}$ to learn the MHI representation from the reference image $I_r$ with size of $H\times W\times 3$, 
\begin{equation}
    C(u_r,v_r,\bm{n},d), \alpha(u_r,v_r,\bm{n},d)\!=\! \phi_{R}(V_{H}(I_r), V_{H}(I_s)).
    \label{eq:perception_net}
\end{equation}
Specifically, we first use planar homography \cite{hartley2003multiple} which is formulated by the camera parameters of input images and layered plane parameters $(\bm{n}^\top, d)^\top$ to establish the \textit{Oriented Cost Volume} (OCV) $V_H(I_r)$ and $V_H(I_s)$ as the inputs of an MHI reconstruction module. 
We then concatenate the layers of $V_H(I_r)$ and $V_H(I_s)$ with the same normal and pass them into the MHI reconstruction module, as shown in Fig. \ref{fig:network}. This concatenated OCV (of the size $2N\!D \times H \times W \times 3$) allows the MHI reconstruction module to better perceive the scene geometry by comparing $I_r$ to each planar homography of $I_s$. Similar to \cite{mildenhall2019local}, the proposed module could additionally learn a blending weight between $V_H(I_r)$ and $V_H(I_s)$, but we found it is easy to lead to ghost artifacts and destroy scene structure. 
In addition, a straightforward choice of the module is employing a single 3D U-Net to perform the whole OCV and predict MHIs directly. However, with the increasing number of discrete normals and distances, the module is too complex to train, especially combining the normal-blending module together. 

\begin{figure}[tb]
    \centering
    \includegraphics[width=\hsize]{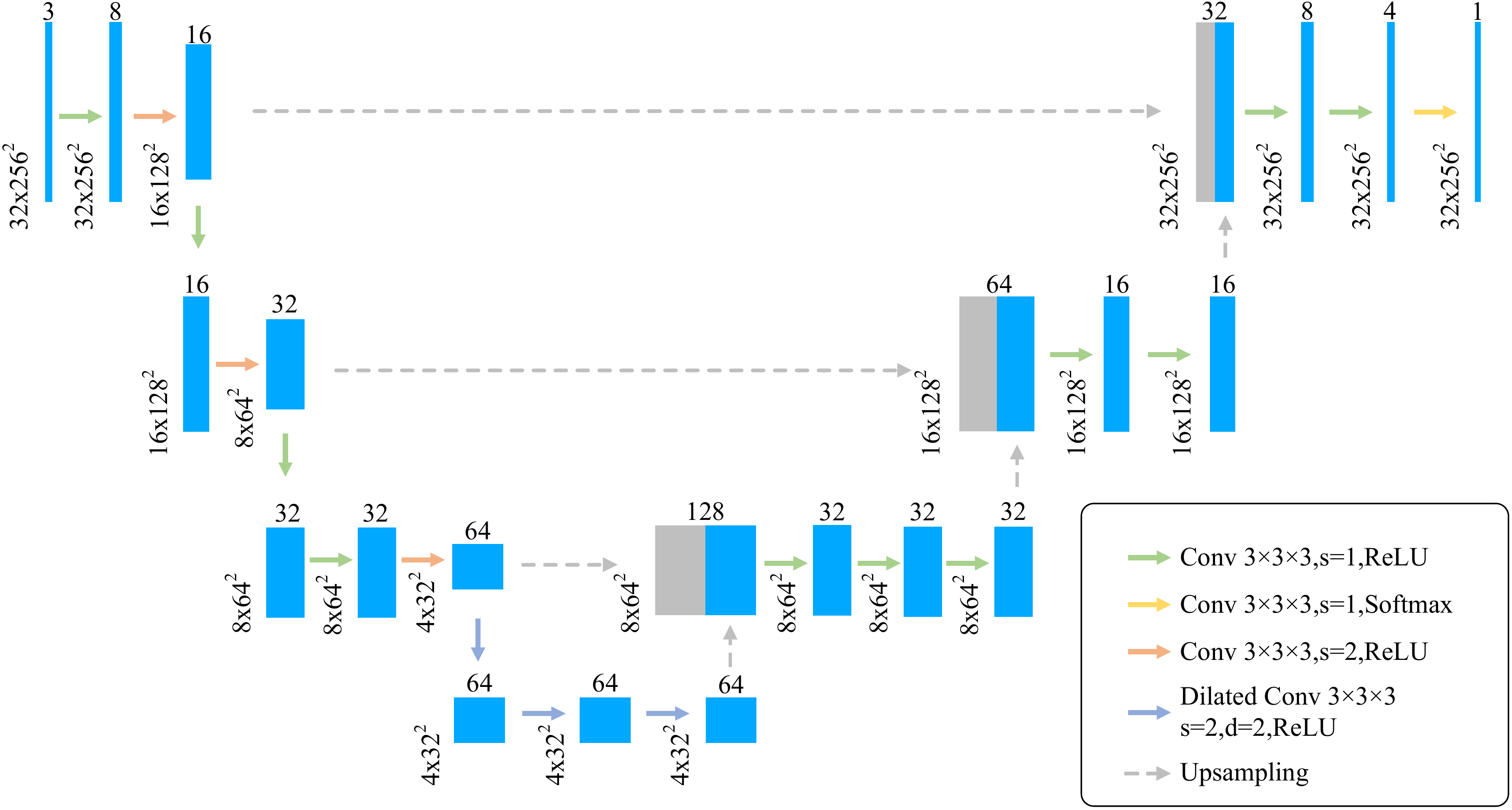}
    \caption{The network architecture of the MHI reconstruction module, where s is the stride, d the kernel dilation. ``Upsampling" donates $2\times$ nearest neighbor upsampling. }
    \label{fig:MHI}
\end{figure}

In practice, when the OCV tensor is too huge for the MHI reconstruction module, we divide the concatenated OCV into several volumes according to the normal and pass them into different 3D U-Nets. It is interesting to note that layers of OCV with the same normal are equivalent to a plane-sweep volume with specific adjustments. It also handles the novel view synthesis independently, similar to MPI. Consequently, to speed up the training processing, we first use the plane-sweep volume with a specific normal to pre-train a 3D U-Net. These pre-trained 3D U-Nets for different discrete normals then form the MHI reconstruction module to perform OCV. Finally, we obtain MHIs with the size of $N\!D \times H \times W \times 4$, namely a set of RGBA images, where the color image records the scene plane and the visible map encodes the opacity and transparency.

Given the MHI representation with respect to the reference camera, each layered plane can be mapped to the target camera via differentiable homography (Eq. \ref{eq:diff_homography}), followed by an alpha composition of the parallel MHIs to a multi-normal image $\hat{C}_i$ in a back-to-front order. In addition, the bilinear interpolation is used to resample sub-pixels.

\textit{Network architecture.} Fig. \ref{fig:MHI} demonstrates a detailed specification of a fully-convolutional encoder-decoder architecture, which follows a similar design as 3D U-Net \cite{cciccek20163d}. The input of MHI reconstruction module is the Oriented Cost Volume (OCV) which is estimated by two images and fixed homography of each layer. Dilated convolutions \cite{chen2017deeplab, yu2015multi} are used to perceive global scene context 
and maintain the spatial resolution of the feature maps, while skip-connections are used to capture the lower scene feature. All layers except the last are followed by a $\mathsf{ReLU}$ nonlinearity and layer normalization \cite{ba2016layer}. The outputs of the final layer are just passed through a $\mathsf{ReLU}$ to reconstruct the visible maps of MHIs (a set of RGBA images consisted of color images and visible maps), while the reference image is copied as the color images.

\subsection{Normal-Blending Module}\label{subsec:network}
Deep learning has been demonstrated to perform image-based tasks and is thus an ideal candidate for solving blending problem of multi-normal images. We train a deep neural module to learn normal-dependent blending weights that are then used to combine warped contributions from multi-normal images, where the goal is to make the blending image look as natural as possible. 

Considering that multi-normal images may share the same content but using parallel planes may discretize scenes in different directions, as discussed in Sec. \ref{subsec:normal_blending}, it poses strict requirements in terms of network architecture, and precludes involved solutions such as Recurrent Neural Network \cite{gregor2015draw}. In addition, we need a convolutional neural module that is able to achieve pixel-wise blending from multi-normal images in a multi-scale fashion closest to scene representation.
For these reasons, we utilize a 2D U-Net-based \cite{ronneberger2015u} convolutional architecture to predict blending weights $\{\hat{w}_i\}_{i=1}^{N}$, \textit{i.e.} normal-blending module $\phi_{B}$,
\begin{equation}
    \{\hat{w}_i\}=\phi_{B}\left((\hat{C}_1,\omega_1), \cdots, (\hat{C}_N,\omega_N)\right),
    \label{eq:blending_net}
\end{equation}
where $\{\hat{C}_i\}_{i=1}^{N}$ and $\{\omega_i\}_{i=1}^{N}$ indicate multi-normal images (see Sec. \ref{subsec:MNI}) and hand-crafted heuristic blending weights (see Sec. \ref{subsec:normal_blending}) respectively.

In details, we use an angle-based cost (Eq. \ref{eq:angle_cost}) to calculate a hand-crafted heuristic blending weight according to the hard scheme or soft scheme for each multi-normal image. It is easy for the hand-crafted heuristic schemes to lead ghost or seam during normal blending, but the computed weights also contain the information of scene geometry. Each multi-normal image $\hat{C}_i$ with the size of $H \times W \times 3$ is concatenated with a blending weight $\omega_i$, and all of them result in a $H \times W \times 4N$ tensor that passed into a normal-blending module, as shown in  Fig. \ref{fig:network}. The outputs are transferred to a softmax to render the target image $\hat{I}_t$ according to Eq.~\ref{eq:MNI_blending}. 

\begin{figure}[tb]
    \centering
    \includegraphics[width=\hsize]{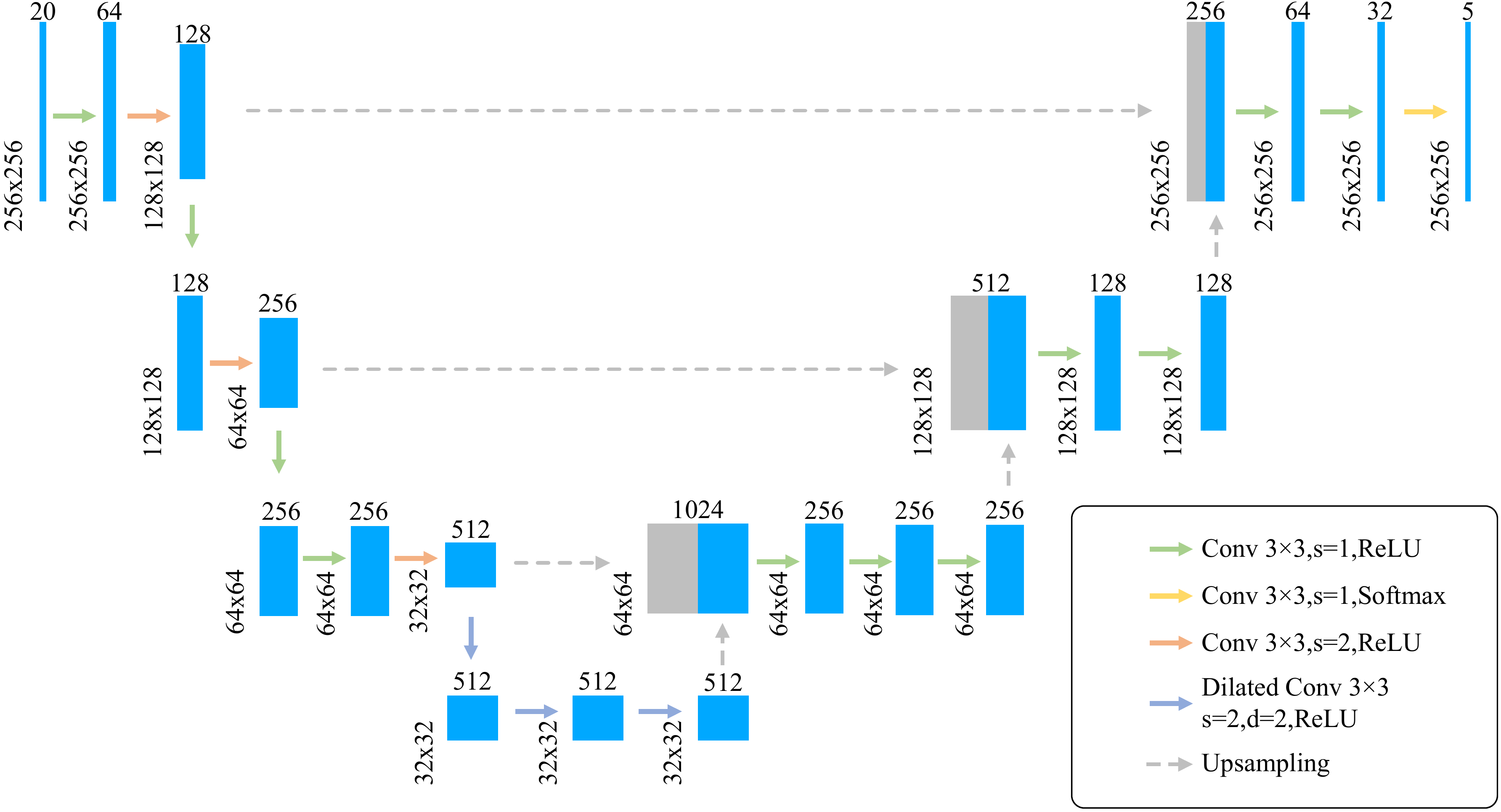}
    \caption{The network architecture of the normal-blending module, where s is the stride, d the kernel dilation. ``Upsampling" donates $2\times$ nearest neighbor upsampling.}
    \label{fig:normal}
\end{figure}

\textit{Network architecture.} We use a 2D U-Net-based \cite{ronneberger2015u} fully-convolutional encoder-decoder architecture, as shown in Fig. \ref{fig:normal} for detailed specification. The inputs of normal-blending module are multi-normal images, each of which is rendered by MHIs along same normal direction. Similar to MHI reconstruction module, the normal-blending module also uses the architecture with dilated convolutions \cite{chen2017deeplab, yu2015multi} and skip-connections. Each Layer is followed by a $\mathsf{ReLU}$ nonlinearity and layer normalization \cite{ba2016layer} except for the last layer, where $\mathsf{Softmax}$ is used and no layer normalization is applied.
Each of the last layer outputs ($N$ blending weights of multi-normal images) is further scaled to match the corresponding valid range, namely $[0, 1]$.

\subsection{Loss Function and Training Scheme}\label{subsec:loss_scheme}

To train the MHI reconstruction module and normal-blending module, we use the synthesized views as supervision. For training loss, we utilize the perceptual loss of \cite{chen2017photographic}.  Given the synthesized image $\hat{I}_t$ and the ground-truth image $I_t$ in the target view, the loss function is 
\begin{equation}
    \mathcal{L}(\hat{I}_t, I_t)=\left\|\hat{I}_t - I_t\right\|_1+\sum_{l}{\lambda\left\|\phi_l(\hat{I}_t)-\phi_l(I_t)\right\|_1},
    \label{eq:vgg_loss}
\end{equation}
where $\{\phi_l\}_{l=1}^{5}$ is the conventional neural layers `$\mathsf{conv1\_2}$', `$\mathsf{conv2\_2}$', `$\mathsf{conv3\_2}$', `$\mathsf{conv4\_2}$' and `$\mathsf{conv 5\_2}$' of a pre-trained VGG-19 network \cite{simonyan2014very}. The weighting coefficients $\{\lambda_l\}_{l=1}^5$ are the same as \cite{chen2017photographic}.


\begin{table*}[t]
    \centering
    \small
    \caption{Quantitative comparisons with baseline method \cite{zhou2018stereo}  (top) and ablation studies (middle) across a wide range of camera rotations
    (measured by the maximum angle of relative rotations $\theta_{\text{max}}$).
    }
    \begin{tabularx}{\textwidth}{lCCCCCCCCC}
        \toprule
        & \multicolumn{3}{c}{$\theta_{\text{max}}\leq2^\circ$} & \multicolumn{3}{c}{$2^\circ<\theta_{\text{max}}\leq4^\circ$} & \multicolumn{3}{c}{$4^\circ<\theta_{\text{max}}\leq8^\circ$} \\
        \cmidrule(lr){2-4}\cmidrule(lr){5-7}\cmidrule(lr){8-10}
        & PSNR $\uparrow$ & SSIM $\uparrow$ & LPIPS $\downarrow$ & PSNR $\uparrow$ & SSIM $\uparrow$ & LPIPS $\downarrow$ & PSNR $\uparrow$ & SSIM $\uparrow$ & LPIPS $\downarrow$\\
        \midrule
        MPI ($D$=32) & 29.58 & 0.936 & 0.061 & 26.03 & 0.923 & 0.069 & 24.39 & 0.906 & 0.080 \\
        MPI ($D$=160) & 29.67 & 0.935 & 0.058 & 26.28 & 0.923 & 0.067 & 24.46 & 0.907 & 0.078 \\
        Ours ($N$=1, $D$=160) & 29.42 & 0.939 & 0.060 & 25.65 & 0.922 & 0.070 & 24.12 & 0.908 & 0.081 \\
        \midrule
        Avg.  & 28.64 & 0.927 & 0.087 & 25.61 & 0.912 & 0.096 & 24.34 & 0.895 & 0.108  \\
        Hard & 30.06 & 0.939 & 0.060 & 27.65 & 0.929 & 0.064 & 26.39 & 0.912 & 0.077\\
        Soft & 30.20 & 0.942 & 0.064 & 27.54 & 0.928 & 0.070 & 26.26 & 0.911 & 0.084\\
        \midrule
        Ours ($N$=5, $D$=32) & \textbf{30.51} & \textbf{0.943} & \textbf{0.054} &\textbf{ 27.96} & \textbf{0.932} & \textbf{0.060} & \textbf{26.59} & \textbf{0.916} & \textbf{0.072}  \\
        \bottomrule
    \end{tabularx}
    \label{tab:comparison_ablation_rot}
\end{table*}

The training scheme of the proposed two-stage network consists of two phases based on the loss function (Eq. \ref{eq:vgg_loss}). In the first phase, we train the MHI reconstruction module from scratch with the loss function defined by multi-normal images $\{\hat{C}_i\}$ and the ground truth image $I_t$ in the target view. In practice, different modules with the same 3D U-Net-based architecture are utilized to perceive the MHIs. Each module is specifically trained for the MHIs with constant normal, so we train them in different modules separately based on the loss function $\mathcal{L}(\hat{C}_i, I_t)$. In the second phase, once the MHI reconstruction module is trained, we concatenate the pre-trained module with the normal-blending module and then learn the normal blending weights $\{\hat{\omega}_i\}$ via $\mathcal{L}(\sum{\hat{\omega}_i\hat{C}_i}, I_t)$ for view synthesis together.

\section{Experiments}

\subsection{Implementation Details}
We train and evaluate on the open-source YouTube RealEstate 10K dataset \cite{zhou2018stereo}.
We randomly select 2,500 video clips as our training dataset and an additional 500 video clips for testing, where the test video clips do not overlap with those in the training dataset. During the training and testing, we generate examples by sampling two source frames and a target frame from a randomly chosen video clip so that the half of the target images are view extrapolation and the other half are view interpolation.

\begin{figure*}[tb]
    \centering
    \includegraphics[width=\linewidth]{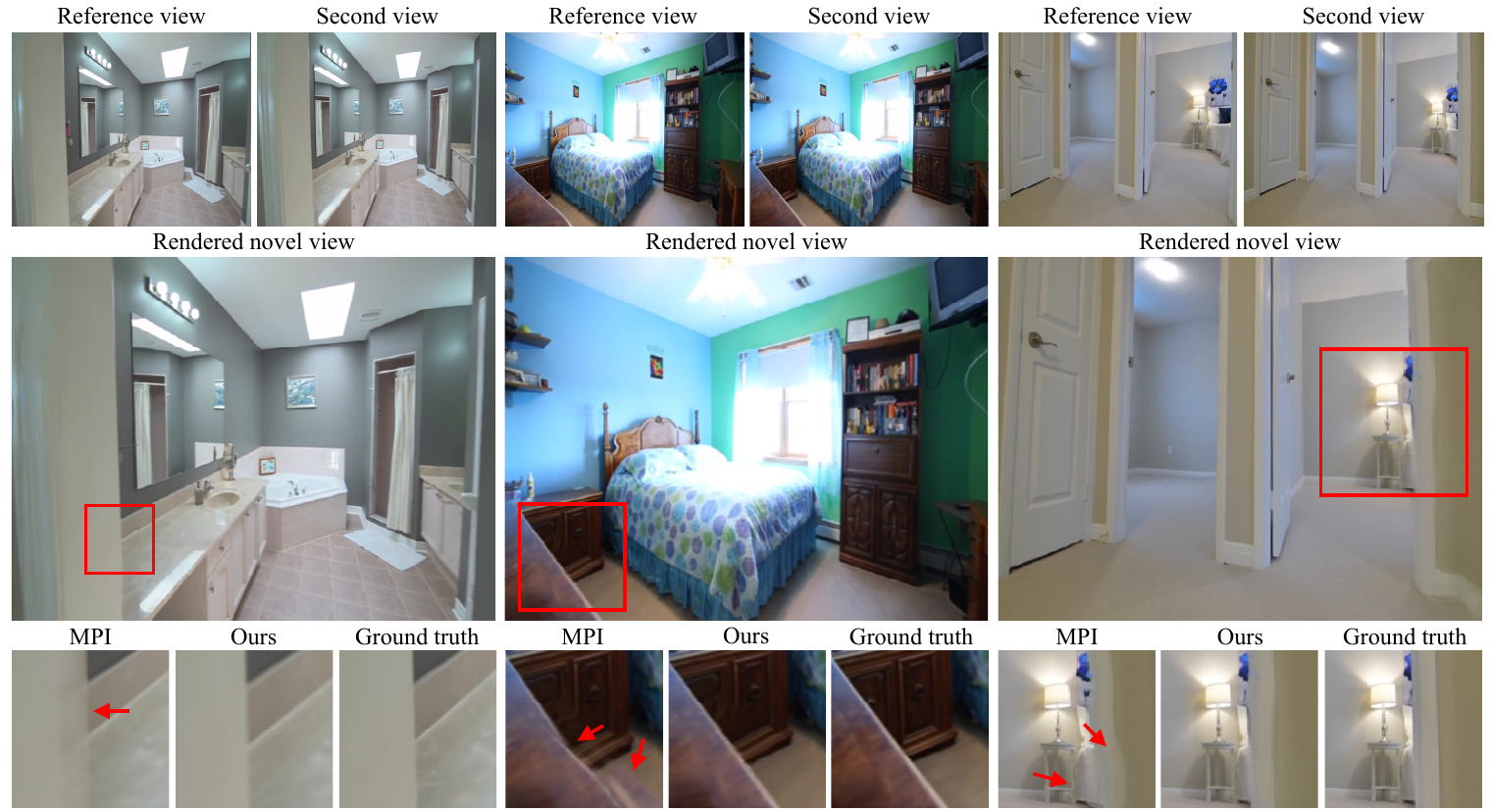}
    \caption{Qualitative comparison of rendered novel view. Compared with MPIs \cite{zhou2018stereo} and ground truth, the reconstructed MHIs find approximate normal for the scene surface instead of the camera-parallel plane to avoid the ghosting artifacts, such as the wall (left image) and the door frame (right image). MPI may divide the scene surface with a large slope into different depth layers and break the structure of thin edges, as demonstrated by the chair in the right image and the table in the middle image.}
    \label{fig:comparison}
\end{figure*}

Unless specified statements, the number of normals and distances is set to $N\!=\!5$ and $D\!=\!32$. Considering the horizon and vertical fields of view are $90^\circ$ and $60^\circ$, we set normals of the layered planes as $\bm{n}_1\!=\!(\frac{\sqrt{2}}{2},0,\frac{\sqrt{2}}{2})^\top$, $\bm{n}_2\!=\!(0,-\frac{\sqrt{3}}{2},\frac{1}{2})^\top$, $\bm{n}_3\!=\!(0,0,1)^\top$, $\bm{n}_4\!=\!(0,\frac{\sqrt{3}}{2},\frac{1}{2})^\top$, $\bm{n}_5\!=\!(-\frac{\sqrt{2}}{2},0,\frac{\sqrt{2}}{2})^\top$. The distance of layered planes is computed by depth from $1m$ to $100m$ which is linearly sampled in disparity. 


We follow the training scheme described in Sec. \ref{subsec:loss_scheme}. We use the novel view as supervision to first train the MHI reconstruction module for $380k$ iterations, then freeze the MHI reconstruction module and train the normal-blending module for $350k$ iterations. The size of input images and MHIs for the proposed network is set as $256\times256$ for training, due to GPU memory limitations. The optimization we used for training is Adam solver \cite{kingma2014adam} with a learning rate of $2\times 10^{-4}$ and a batch size of one. The proposed network is implemented on the Tensorflow \cite{abadi2016tensorflow}. Training takes about 4 days on a single Titan-X GPU device.

Our method is compared with the baseline MPI reconstruction provided by Zhou \textit{et al.} \cite{zhou2018stereo}. The reasons are that: it outperforms other recent view synthesis methods for light-field rendering \cite{zhang2015light, kalantari2016learning}, and its results are impressive but reflect the common problems (\textit{e.g.} a scene plane belongs to different layers and structure inconsistency) in depth-dependent representations \cite{srinivasan2019pushing,choi2019extreme}. We implement the baseline from the original code provided by Zhou \textit{et al.} \cite{zhou2018stereo}. The same training set of our algorithm is used to train MPI, which is supervised by the novel view. It is necessary to note that the number of layers used in MPI is set to $160$ for the fair comparison. The optimization we used for training is the same to the setting of our method with $500k$ iterations. 

We measure the quality of the synthesized images using the following metrics: PSNR, SSIM~\cite{wang2004image}, and LPIPS \cite{zhang2018unreasonable}. Here, a higher PSNR and SSIM and a lower LPIPS all indicate better results. 
Besides, we use rotation angle in degree $\theta\!=\!\arccos({{(\mathsf{trace}(\bm{R})-1)}/{2}})$ to measure camera rotation.


\begin{figure*}[tb]
    \centering
    \includegraphics[width=\hsize]{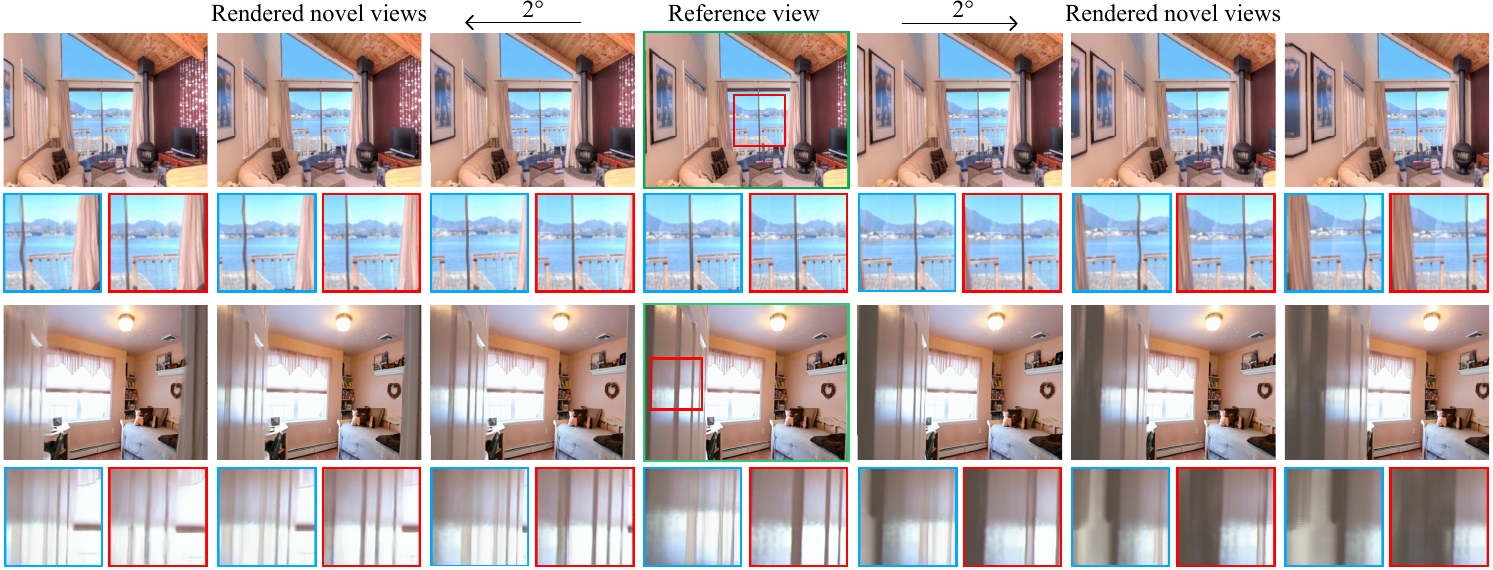}
    \caption{Circular light-field rendering using our method. The MHIs of the reference image (\textcolor{myGreen}{green}) are inferred by the MHI reconstruction module. The inferred MHIs are used to extrapolate and interpolate views forward and backward rotated with a step of $2^\circ$ along a circle (radius of $2m$). Our method (\textcolor{myRed}{red} in the right) performs better than MPI (\textcolor{myBlue}{blue} in the left). The results in the second and fourth columns demonstrate our method preserves the global structure and avoids ghosting artifacts respectively, wherein renderings of the reference view are directly synthesized from inferred MHIs without homography. }
    \label{fig:view_rotation}
\end{figure*}

\begin{figure*}[tb]
    \begin{minipage}[b]{0.205\textwidth}
        \centering
        \subfigure{
            \centering
            \setlength{\fboxsep}{0pt}%
            \setlength{\fboxrule}{1pt}%
            \fcolorbox{myGreen}{white}{\includegraphics[width=\textwidth]{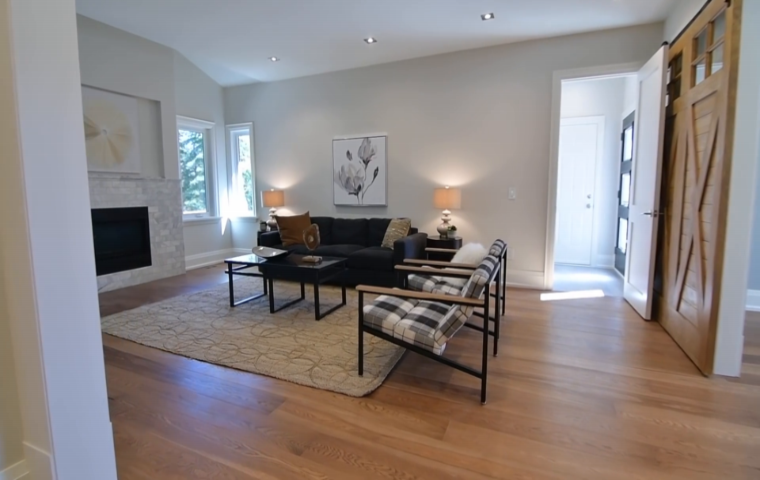}}
        }
        \subfigure{
            \centering
            \setlength{\fboxsep}{0pt}%
            \setlength{\fboxrule}{1pt}%
            \fcolorbox{black}{white}{\includegraphics[width=\linewidth]{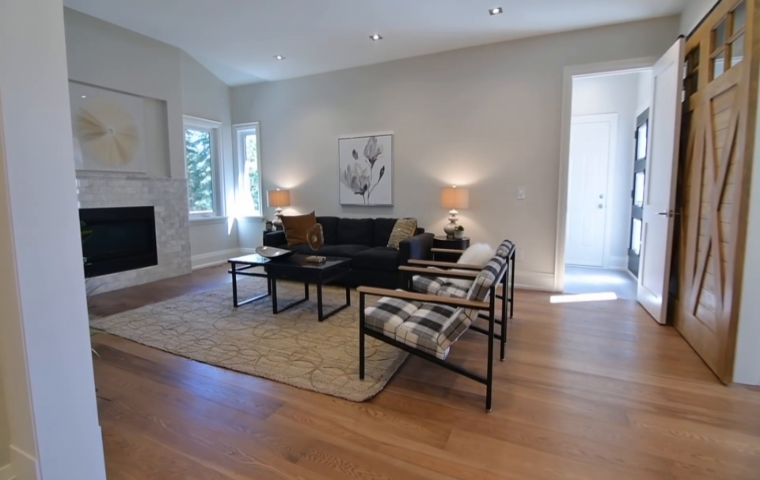}}
        }
    \end{minipage}\quad
    \begin{minipage}[b]{.315\textwidth}
        \centering
        \subfigure{
            \centering
            \animategraphics[autoplay,loop, width=.465\linewidth]{3}{mhi/mhi_0_}{0}{15}
            }\hspace{-1mm}
        \subfigure{
            \centering
            \animategraphics[autoplay,loop, width=.465\linewidth]{3}{mhi/mhi_4_}{0}{15}
        }\vspace{-1mm}
        \subfigure{
            \centering
            \animategraphics[autoplay,loop, width=.47\linewidth]{3}{mhi/mhi_2_}{0}{15}
        }\vspace{-1mm}
        \centering
        \subfigure{
            \centering
            \animategraphics[autoplay,loop, width=.465\linewidth]{3}{mhi/mhi_1_}{0}{15}
        }\hspace{-1mm}
        \subfigure{
            \centering
            \animategraphics[autoplay,loop, width=.465\linewidth]{3}{mhi/mhi_3_}{0}{15}
        }
    \end{minipage}\hspace{-0.2mm}
    \begin{minipage}[t]{.45\textwidth}
        \centering
        \animategraphics[autoplay,loop, width=\linewidth]{10}{mhi/out_}{0}{40}
    \end{minipage}
    \put(-480,-10){{\scriptsize (a) Input images}}
    \put(-360,-10){{\scriptsize (b) Inferred MHI representation}}
    \put(-173,-10){{\scriptsize (c) Novel views synthesized from MHI}}
    \put(-365, 144){\scriptsize \colorbox{white}{$\bm{n}_1$}}
    \put(-280, 144){\scriptsize \colorbox{white}{$\bm{n}_5$}}
    \put(-322, 94){\scriptsize \colorbox{white}{$\bm{n}_3$}}
    \put(-365, 44){\scriptsize \colorbox{white}{$\bm{n}_2$}}
    \put(-280, 44){\scriptsize \colorbox{white}{$\bm{n}_4$}}
    \caption{Visualization of the proposed method, including (a) the input image pair, (b) our inferred MHI representation of the reference image (\textcolor{myGreen}{green}), which we show the alpha-multiplied videos from near to far with different normals, and (c) novel videos rendered from the MHI. The predicted MHI is able to capture the scene appearance and geometry information in a layer-wise manner (near to far). \textbf{Please note that this figure contains video clips}. Should this figure not already be playing then please consider viewing this paper using Adobe Reader.}
    \label{fig:v_mhi}
\end{figure*}

\begin{figure*}[t]
    \begin{minipage}[t]{.675\textwidth}
        \centering
        \includegraphics[width=\linewidth]{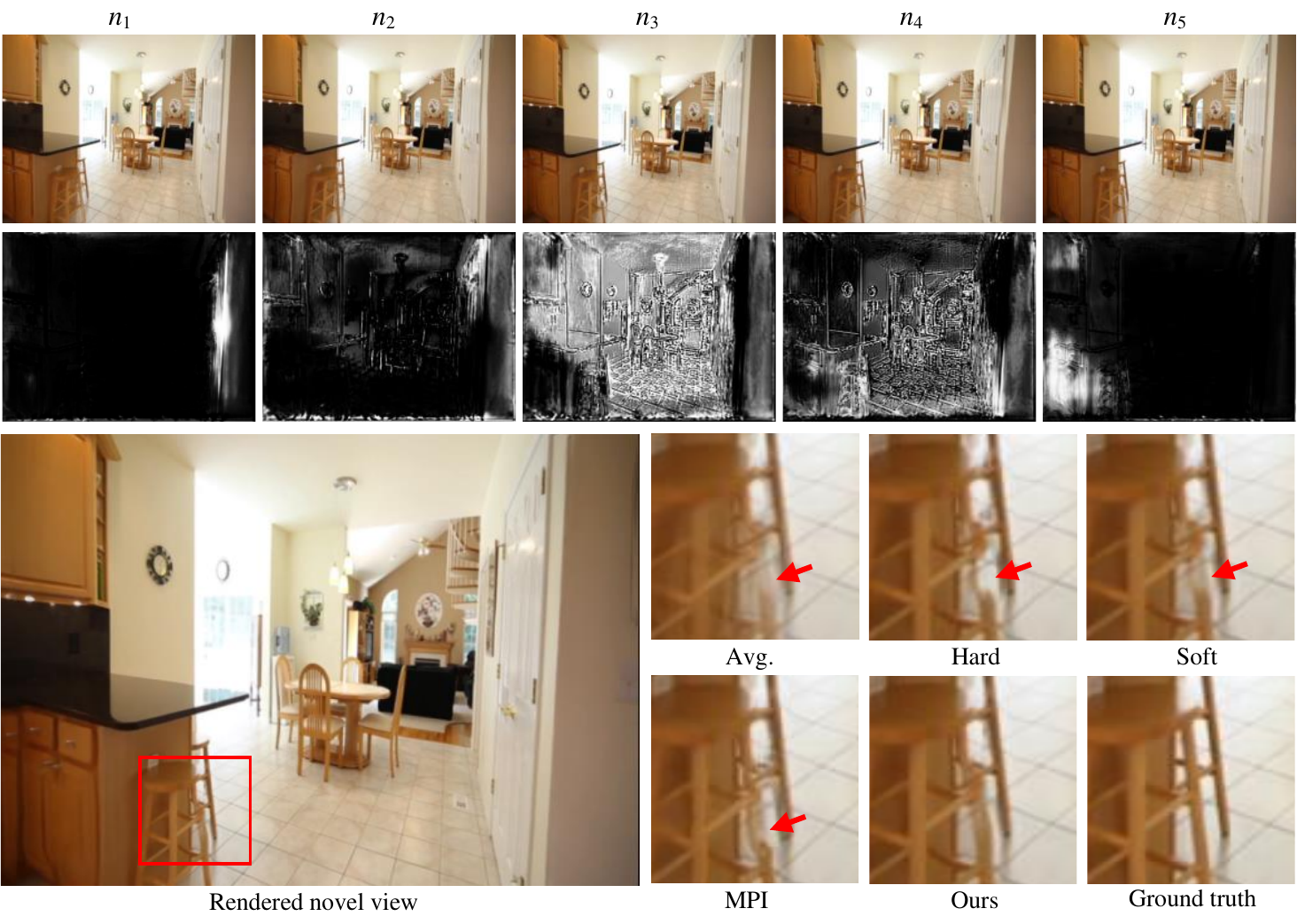}
        \caption{Ablation study on different blending schemes of multi-normal images. Multi-normal images (first row) are projected and composited from MHIs along different normals. Normal blending weight (second row) for each multi-normal image is learned via a normal-blending module.}
        \label{fig:ablation}
    \end{minipage}\quad
    \begin{minipage}[t]{.295\textwidth}
        \centering
        \includegraphics[width=\linewidth]{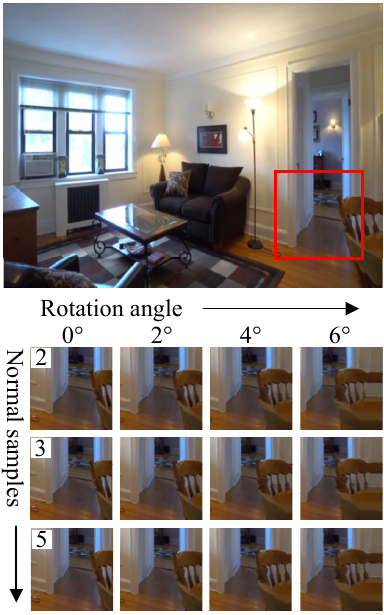}
        \caption{Effect of varying normal samples at different rotation angles. We show details of extrapolated unstructured views (zooming in). }
        \label{fig:ablation_rot}
    \end{minipage} 
\end{figure*}

\subsection{Comparison with Baseline Method} 
We quantitatively (Tab. \ref{tab:comparison_ablation_rot}) and qualitatively (Fig. \ref{fig:comparison}) demonstrate that the proposed method produces superior renderings, particularly for the cases with camera rotations and the scenes with slanted planes, without the artifacts seen in renderings from competing method. We urge readers to view our accompanying demo video that highlights the benefits of the proposed method.

First quantitative comparisons are shown in Tab. \ref{tab:comparison_ablation_rot} from which our method outperforms the MPI across wide ranges of camera rotations. Note that the camera translation is random within the disparity of $64$ pixels similar to \cite{zhou2018stereo}. Besides, we follow the training scheme described in Sec. \ref{subsec:loss_scheme} to implement our algorithm with the normal of $\bm{n}_3$ and $D\!=\!160$ distance layers. The results of our method ($N=1,D=160$) are similar to that of MPI, as shown in Tab. \ref{tab:comparison_ablation_rot}. It verifies that the superior performance of our method is caused by introducing normal samples instead of increasing distance samples or modifying network architecture. The 3D U-Net-based architecture is used for MHI reconstruction due to the vast and complex 3D OCV.
In particular, we improve the average PSNR scores of MPI by $2.83\%$, $6.39\%$ and $8.71\%$ with the increasing rotation angles. 

The experiments of MPI with different depth layers are also conducted, which are labeled by MPI ($D\!=\!32$) and MPI ($D\!=\!160$) in Tab. \ref{tab:comparison_ablation_rot} respectively. It demonstrates that the performance of the baseline method has little improvement with the increasing layers. The reason for similar performance is that the depth sampling has an upper bound. With the increasing depth layers of MPI, it is easy to assign the color image and opacity to incorrect layers in regions of ambiguous or repetitive texture and regions, especially on the slanted surface. Furthermore, the comparisons of our algorithm with same samples but different setting are performed. The distance sampling has an upper bound, while normal sampling improves scene representation performance even more. It verifies that our method outperforms the MPI due to the introduction of normal sampling rather than the design of network.

Moreover, we qualitatively compare our method with both ground truth and MPI in Fig. \ref{fig:comparison}. It demonstrates that our method produces superior renderings, especially for structure consistency and avoiding ghosting and other abnormalities  on slanted surfaces.
We also implement an application of our trained model on circular light-field rendering with orbiting motion, \textit{i.e.} a set of rotated views around the center of the scene. 
The results shown in Fig. \ref{fig:view_rotation} verify plausible interpolations and extrapolations of our method, in particular for the spatial structure consistency. Besides, temporal artifacts are imperceptible with orbiting motion, also shown in the demo video.

Finally, we visualize examples of the MHI representation inferred by the proposed two-stage network in Fig. \ref{fig:v_mhi}. Despite the lack of a direct color or alpha ground-truth for each MHI plane during training, the inferred MHI is able to record scene appearance and geometry information layer-by-layer (near to far, with varied normals), allowing realistic rendering of novel views from the representation.



\subsection{Ablation Study}
\subsubsection{Blending Scheme}
We validate the proposed hand-crafted heuristic blending schemes of multi-normal images (Sec. \ref{subsec:normal_blending}) as well as the proposed normal-blending module (Sec. \ref{subsec:network}) in Tab. \ref{tab:comparison_ablation_rot}. We first project and render the multi-normal images in the target view according to the MHIs predicted from the MHI reconstruction module. We then perform different blending schemes for unstructured view synthesis, where `Avg.' indicates a simple average of multi-normal images. `Hard' and `Soft' refer to the renderings from multi-normal images using hard and soft blending schemes as shown in Fig. \ref{fig:MNI_blending}(c) respectively. 
Fig. \ref{fig:ablation} illustrates the benefits of using normal-blending module to blend multi-normal images. The blending weights for each normal exactly reflect the normal distribution of the scene, as shown in the second row of Fig. \ref{fig:ablation}. Specifically, the normals of the chair and table at left-bottom of Fig. \ref{fig:ablation} are closer to $\bm{n}_5$ rather than $\bm{n}_3$ (camera-parallel normal). Similarly, the right wall and the floor are close to $\bm{n}_1$ and $\bm{n}_4$ respectively. It is the reason that our method outperforms the baseline method. 
In addition, compared with the normal-blending module, average and soft blending schemes softly compute the weights and tend to generate ghosting and aliasing artifacts, while the hard blending scheme breaks the global structure, as shown in Fig. \ref{fig:ablation}.

\begin{table}[tb]
    \centering
    \small
    \caption{Evaluation of varying numbers of normals and distances for the MHI representation.}
    \begin{tabularx}{0.475\textwidth}{ccCCC}
        \toprule
        $N$ & $D$ & PSNR $\uparrow$ & SSIM $\uparrow$ & LPIPS $\downarrow$ \\
        \midrule
        \multirow{3}{*}{2} & 8 & 24.93 & 0.835 & 0.138 \\
        & 16  & 26.74 & 0.881 & 0.104 \\
        & 32  & 29.21 & 0.928 & 0.062\\
        \midrule
        \multirow{3}{*}{3} & 8 & 26.10 & 0.866 & 0.112 \\
        & 16 & 27.39 & 0.893 & 0.087 \\
        & 32 & 29.33 & 0.929 & 0.060 \\
        \midrule
        \multirow{3}{*}{5} & 8 & 26.05 & 0.863 & 0.114 \\
        & 16 & 27.88 & 0.901 & 0.085 \\
        & 32 & 29.51 & 0.932 & 0.059 \\
        \bottomrule
    \end{tabularx}
    \label{tab:evaluation_layers}
\end{table}


\begin{figure}[tb]
    \centering
    \includegraphics[width=\hsize]{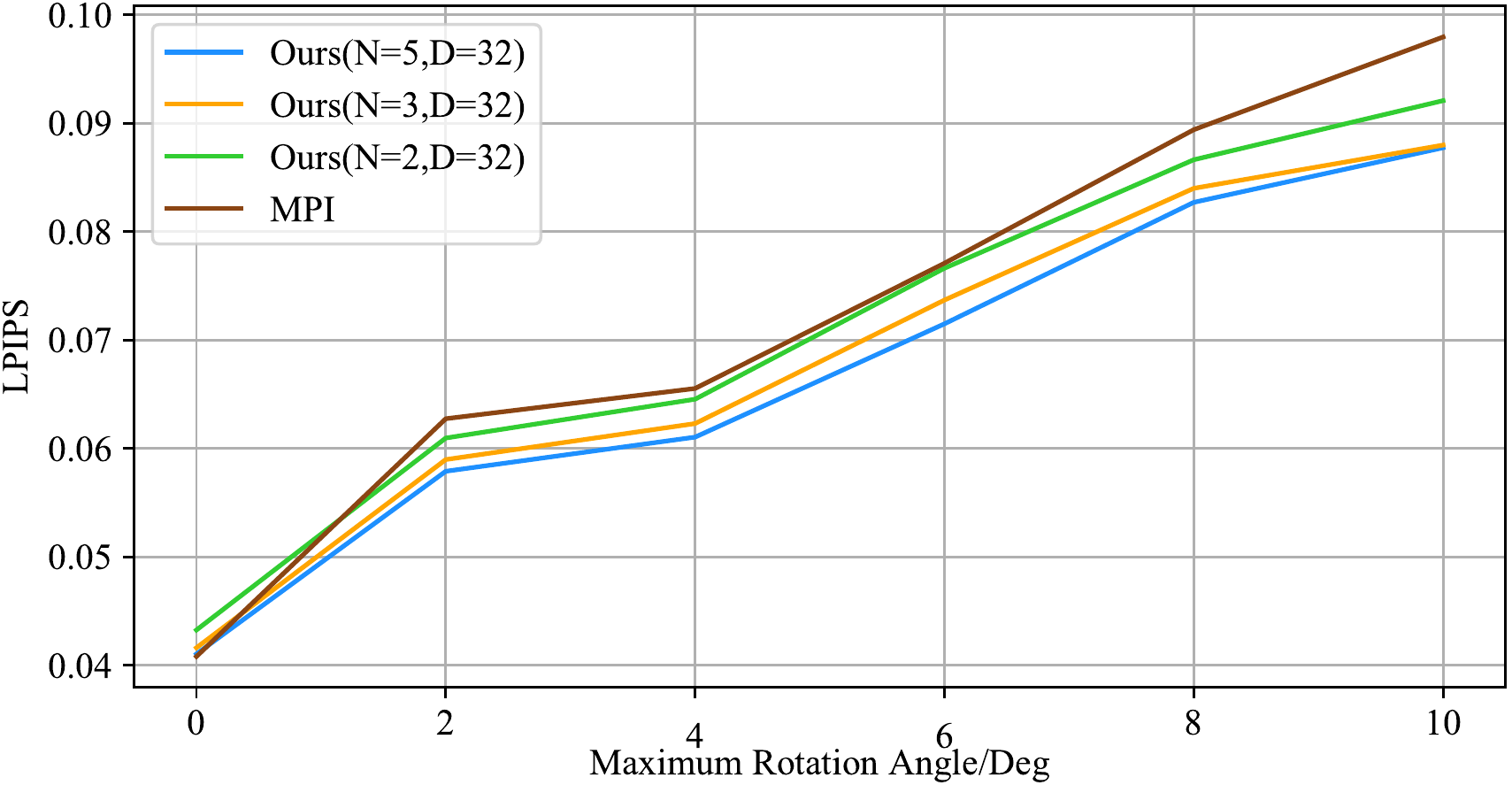}
    \caption{LPIPS comparisons with MPI \cite{zhou2018stereo} for different maximum rotation angles. Our method outperforms MPI and the gap increases with maximum rotation angle.}
    \label{fig:rot_comp}
\end{figure}

\subsubsection{Number of Samples}
We first investigate the influence of the number of normals and distances on the effectiveness of our method in Tab. \ref{tab:evaluation_layers}, where the normal samples of $N\!=\!2$ and $N\!=\!3$ are $\{\bm{n}_1, \bm{n}_5\}$ and $\{\bm{n}_1, \bm{n}_3, \bm{n}_5\}$ respectively. As shown in Tab. \ref{tab:evaluation_layers}, we can see that the performance of the proposed method improves with the number of layered planes significantly, except for the case with varying normal layers and fixed distance layers to $D\!=\!8$. The reason for this phenomenon is that the sampling of distance is so sparse that the sparsely sampled normals do not work. Moreover, we plot the performance (evaluated by LPIPS) of our method (with varying normal samples $N\!=\!2,3,5$) compared with MPI \cite{zhou2018stereo} for different camera rotations in Fig. \ref{fig:rot_comp}. Our method can achieve a similar perceptual quality as MPI with a slight rotation, but our method performs better and the difference widens with increasing angle. We use the LPIPS metric (lower is better) because we primarily value perceptual quality which is noticeable to the human eye \cite{dosselmann2011comprehensive}. Fig. \ref{fig:ablation_rot} also shows view extrapolations from MHIs of different normals that rotates along a circle (radius of $2m$) with a step of $2^\circ$. It is noting that the larger normal samples, the farther the view can be rotated before introducing artifacts, \textit{e.g.} edges of the chair.

\subsection{Limitations}
A primary limitation of our algorithm is the vast cost volume of the MHI reconstruction module, which increases the time complexity and difficulty of network convergence. 
The normal-blending module also has to estimate blending weights for each novel view, which is also a limitation. It remains a challenge to synthesize views in real time. However, with the introduction of normal sampling, occlusions on the target image are more delicate compared with the camera frontal parallel representation. Thus a normal-blending module not only searches the corresponding MHI for each part of the scene but avoids occlusions.

\section{Closing Remarks}
View synthesis has gained increasing attention and has been used to many application including light-field rendering for immersive visual experience. Although the popular MPI representation models the complex appearance effects, it suffers from ghosting as well as other artifacts, especially for scene planes with significant slopes, and it is incapable of handling unstructured views. We have proposed a compact MHI representation, which discretizes the scene as planes at a fixed range of normals and distances. Besides, by analyzing per-normal geometry, an angle-based cost is proposed to guide the blending process of the multi-normal images. A two-stage network is proposed to take advantage of the MHI representation and per-normal geometry, which achieves superior performance than the state-of-the-art method.

\ifCLASSOPTIONcaptionsoff
  \newpage
\fi



\bibliographystyle{IEEEtran}
\bibliography{IEEEabrv,egbib}
\end{document}